\documentclass[lettersize,journal]{IEEEtran}

\IEEEoverridecommandlockouts                              

\usepackage{graphics} 

\usepackage{times} 
\usepackage{cite}

\usepackage{amsthm}
\newtheorem*{remark}{Remark}
\usepackage{amssymb}  

\usepackage[linesnumbered,ruled,vlined]{algorithm2e}
\SetKwInput{KwInput}{Input}                
\SetKwInput{KwOutput}{Output}              

\newcommand{\bmu}{{\boldsymbol{\mu}}}

\newcommand{\bSigma}{{\boldsymbol{\Sigma}}}
\newcommand{\calN}{\mathcal{N}}

\newcommand{\bI}{\mathbf{I}}
\newcommand{\bPhi}{\boldsymbol{\Phi}}
\newcommand{\bt}{\mathbf{t}}
\newcommand{\bA}{\mathbf{A}}
\newcommand{\bB}{\mathbf{B}}
\newcommand{\bF}{\mathbf{F}}
\newcommand{\bQ}{\mathbf{Q}}

\newcommand{\bM}{\mathbf{M}}

\newcommand{\bh}{\mathbf{h}}

\newcommand{\bK}{\mathbf{K}}

\newcommand{\bb}{\mathbf{b}}
\newcommand{\bu}{\mathbf{u}}
\newcommand{\bx}{\mathbf{x}}
\newcommand{\bz}{\mathbf{z}}

\newcommand{\bd}{\mathbf{d}}
\newcommand{\bc}{\mathbf{c}}
\newcommand{\bw}{\mathbf{w}}

\newcommand{\balpha}{\boldsymbol{\alpha}}

\newcommand{\mE}{\mathbb{E}}

\usepackage{subcaption} 
\usepackage{color}

\usepackage{mathtools}

\title{
A Gaussian variational inference approach to motion planning
}

\author{Hongzhe Yu and Yongxin Chen
\thanks{Financial support from NSF under grants 1942523, 2008513 are greatly acknowledged.}
\thanks{H. Yu and Y.\ Chen are with the School of Aerospace Engineering,
Georgia Institute of Technology, Atlanta, GA; {\{hyu419,yongchen\}@gatech.edu}}
}

\begin{document}

\maketitle
\thispagestyle{empty}
\pagestyle{empty}

\begin{abstract}
We propose a Gaussian variational inference framework for the motion planning problem. In this framework, motion planning is formulated as an optimization over the distribution of the trajectories to approximate the desired trajectory distribution by a tractable Gaussian distribution. Equivalently, the proposed framework can be viewed as a standard motion planning with an entropy regularization. Thus, the solution obtained is a transition from an optimal deterministic solution to a stochastic one, and the proposed framework can recover the deterministic solution by controlling the level of stochasticity. To solve this optimization, we adopt the natural gradient descent scheme. The sparsity structure of the proposed formulation induced by factorized objective functions is further leveraged to improve the scalability of the algorithm. We evaluate our method on several robot systems in simulated environments, and show that it achieves collision avoidance with smooth trajectories, and meanwhile brings robustness to the deterministic baseline results, especially in challenging environments and tasks.

\end{abstract}



\section{Introduction}



%
%
%
%
Motion planning \cite{latombe2012robot} is a fundamental problem in robotics where the goal is to obtain a sequence of states in the space such that it connects a start and goal state while remaining feasible along the plan. When considering motion planning problems, ubiquitous uncertainties arise from imperfect system modeling and measurement noise. Robust motion planning under uncertainties has attracted attentions in the community. Guaranteed robustness was achieved by control and verification design \cite{tedrake2010lqr, majumdar2017funnel} where uncertainties are implicit in the formulation. Stochasticity can also be explicitly brought into the formulation \cite{kalakrishnan2011stomp}. Planning in belief space \cite{kaelbling2013integrated, van2012motion} models states and measurements as distributions named \textit{`belief'}, and planning and control are conducted in these spaces over distributions. Explicitly encoding stochasticity in motion planning has been shown \cite{kalakrishnan2011stomp} helpful in overcoming locally minimum deterministic solution for non-convex and multimodal \cite{osa2020multimodal} optimization problem.

In this work we propose a Gaussian variational inference (GVI) approach to solve motion planning as a probability inference. \cite{mukadam2018continuous} solved this inference problem using maximum a priori (MAP) estimation. Variational inference (VI) \cite{blei2017variational} used in this paper, on the other hand, approaches inference problems by solving an optimization within a proposed distribution family. Operating on distributions, VI naturally accounts for stochasticity in an explicit way. A natural gradient descent scheme is used to solve the optimization. The linear Gaussian process (GP) representation of the trajectory used in this paper has gained its popularity in planning \cite{mukadam2018continuous} and estimation \cite{barfoot2014batch} since it encodes smoothness and enjoys a sparsity pattern.

Our framework takes into account uncertainties on top of Gaussian Process Motion Planning (GPMP2)\cite{mukadam2018continuous}. We show that the proposed method is equivalently motion planning with entropy regularization. Entropy maximization in motion planning and reinforcement learning have been studied in \cite{lambert2021entropy, ziebart2008maximum} and was shown to increase system's robustness to disturbances \cite{eysenbach2021maximum}. Different from the existing works, our proposed method (1) uses a Newton-style optimization scheme which does not need a sampling scheme or learning process, and (2) is scalable by leveraging the sparsity. (3) The proposed method is shown to be an interpolation from a deterministic solution to a stochastic one. It recovers the deterministic solution by controlling the uncertainty level. (4) We show by experiment that the entropy term encodes the level of risk, which then serves as a metric measuring robustness in decision-making among multiple candidate plans. The optimization scheme for GVI in this paper was first proposed in \cite{opper2009variational}, and has been applied in the robot estimation problems in \cite{barfoot2020exactly}, where the factorized property of the problem was leveraged. To the best knowledge of the authors this is the first work that GVI is used in robot motion planning. 

This paper is organized as follows. Section \ref{sec:related_works} discusses the related works. In Section \ref{sec:problem} we formulate the motion planning problem as variational inference. The method to solve this inference problem is presented in Section \ref{sec:optimization}. Our framework is illustrated in Section \ref{sec:experiments} through numerical experiments.

\section{related work}\label{sec:related_works}
The study of motion planning has a long history in robotics community. Sampling based methods such as Rapidly-exploring random tree (RRT) and Probabilistic road map (PRM) \cite{lavalle1998rapidly} \cite{kavraki1996probabilistic} provide with optimal yet course paths as graphs or trees connecting start and goal configurations. However they do not consider dynamical feasibility of the system in their formulations. Trajectory optimization \cite{ratliff2009chomp, schulman2013finding, schulman2014motion} uses optimal control framework to generate trajectories by formulating the problem as a constrained optimization. Direct or collocation methods \cite{ratliff2009chomp, kalakrishnan2011stomp, mukadam2016gaussian, mukadam2018continuous} operate in control and trajectory space while indirect methods \cite{tassa2014control} optimize only on control inputs, both of which have gained successes in obtaining locally optimal solutions.

The most relevant works to our work is \cite{mukadam2018continuous} and \cite{barfoot2020exactly}. In \cite{mukadam2018continuous} the authors formulated planning as a inference problem and solved it using MAP. \cite{barfoot2020exactly} proposed a sparse Gaussian variational inference method to solve inference problem in robot estimation. Variational inference has been used in motion plannings also in \cite{osa2022motion, osa2020multimodal}. We leverage the connection between motion planning and estimation problems, and we also find interesting connections between the GPMP formulation \cite{mukadam2018continuous}, stochastic control problem, and variational inference problem, as discussed in \cite{CheGeoPav16, chen2016relation}.

Robustness to uncertainties is also one of the main motivations of this formulation. Robust motion planning seeks robustness against environment uncertainties. In \cite{majumdar2017funnel, tedrake2010lqr} the authors compute verifiably safe reachable sets using Lyapunov analysis, where the robustness is measured by the volume of the reachable sets around a nominal trajectory. In this work robustness is encoded in the system entropy which is also proportional to the volume of the covariance matrix in Gaussian case.

\section{Problem formulation}
\label{sec:problem}

In this section we formulate motion planning as a variational inference problem. Our formulation generalizes the Gaussian process motion planning \cite{mukadam2016gaussian} that casts motion planning as a MAP task.

\subsection{Gaussian process motion planning}

Trajectory optimization formulates the motion planning problem as an optimization of the form

\begin{equation}
\begin{split}
    &\min_{\bx(\cdot),\bu(\cdot)} \mathcal{F}(\bx, \bu) \\
    &\; \; {\rm s.t.} \;\; \mathcal{G}_i(\bx, \bu) \leq 0,\; i=1, \dots, m\\
    &\; \; \;\;\;\; \;\; \mathcal{H}_i(\bx, \bu) = 0,\; i=1, \dots, r,
\end{split}
\label{eq:general_trj_opt}
\end{equation}
where $\mathcal{F}$ is the cost function and $\mathcal{G}_i$'s, $\mathcal{H}_i$'s are constraints often related to system dynamics, collision avoidance, or actuation limits. The optimization is over the trajectory $\bx(\cdot)$ and the control input $\bu(\cdot)$ jointly.

The GPMP framework, alternatively, formulates the motion planning as a MAP problem
\begin{equation}
\begin{split}
    \bx^\star &= \underset{\bx}{\arg\max}\; p(\bx|\bz) 
    \\&= \underset{\bx}{\arg\max}\; p(\bz|\bx)p(\bx),
\end{split}
\label{eq:MAP}
\end{equation}
where the prior distribution $p(\bx)$ promotes smoothness of the solution, and the likelihood $p(\bz|\bx)$ of some desired behavior encoded by event $\bz$ enforces collision avoidance. In particular, the prior distribution is associated with a linear Gaussian process 
    \begin{equation}\label{eq:dynamics}
        \dot \bx(t) = \bA(t)\bx(t) + \bF(t)\bw(t)+\bb(t),
    \end{equation}
where $\bw$ denotes standard white noise with covariance $\bQ_c$.

After discretization over time steps $\bt = [t_0, \dots,t_N]$, the trajectory becomes a vector $\bx = [x_0, \dots, x_N]^T$ and the prior becomes a Gaussian distribution $\calN(\bmu, \mathbf{K})$ where the inverse covariance matrix $\mathbf{K}^{-1} = \bB^{T}\bQ^{-1}\bB$ has an interesting sparse structure\cite{barfoot2014batch} with
\begin{equation}
    \bB = \begin{bmatrix}
    \bI    &     &   &   & \\
    -\bPhi(t_1, t_0) & \bI &   &   & \\
     &  &  \dots   &  &\\
     &  &  & -\bPhi(t_N, t_{N-1}) &  \bI\\
     &  &  &  \mathbf{0}& \bI 
\end{bmatrix},
\label{eq:tehta_inv_A}
\end{equation} 
and
\begin{equation}
    \bQ^{-1} = {\rm diag}(\bK_0^{-1}, \bQ_{0,1}^{-1},\ldots, \bQ_{N-1,N}^{-1}, \bK_N^{-1}).
\end{equation}
Here $\bPhi$ is the state transition matrix associated with $\bA(t)$, $\bQ$ is a Grammian defined as $\bQ_{i,i+1} = \int_{t_i}^{t_{i+1}}\bPhi(t_{i+1},s)\bF(s)\bQ_c\bF(s)^T\bPhi(t_{i+1},s)^T ds$, and $\mathbf{K}_0, \mathbf{K}_N$ are desired covariances of the start and goal states.

We note that the likelihood probability $p(\bz|\bx)$ describes in general the probability of the feasibility of the current trajectory candidate in \eqref{eq:general_trj_opt}. In this work we consider collision avoidance likelihood
\begin{equation}\label{eq:likelihood}
    p(\bz|\bx) \propto \exp(-\lVert\bh(\bx)\rVert_{\bSigma_{obs}^{-1}}^{2})
\end{equation}
where $\lVert \bh(\bx) \rVert_{\bSigma_{obs}^{-1}}^2$ is a penalty for the collision constraints. Clearly, the MAP problem \eqref{eq:MAP} is equivalent to minimizing the cost function
    \begin{equation}
    \frac{1}{2} \lVert \bx- \bmu \rVert_{\mathbf{K}^{-1}}^{2} + \lVert \bh(\bx) \rVert_{\bSigma_{obs}^{-1}}^2,
    \label{eq:GPMP2_obj}
\end{equation}
where $\|\cdot\|_{\bK^{-1}}$ denotes weighted 2-norm.

The prior in \eqref{eq:MAP} can be decomposed into factors
\begin{equation}\label{eq:fact_prior}
    p(\bx) \propto f_0(x_0) f_N(x_N) \Pi_{i=0}^{N-1}f_{gp}^i(x_i, x_{i+1})
\end{equation}
with
\begin{eqnarray*}
    f_0(x_0) &=& \exp(-\frac{1}{2}\lVert x_0 - \mu_0 \rVert_{\mathbf{K}_0^{-1}}),\\
    f_N(x_N) &=& \exp(-\frac{1}{2}\lVert x_N - \mu_N \rVert_{\mathbf{K}_N^{-1}}),\\
f_{gp}^i(x_i, x_{i+1}) &=& \exp(-\frac{1}{2}\lVert \bPhi(t_{i+1}, t_i)(x_i-\mu_i) \\&&- (x_{i+1}- \mu_{i+1}) \rVert_{\bQ_{i,i+1}^{-1}}),
\end{eqnarray*}
and the collision cost \eqref{eq:likelihood} can also be factorized into
\begin{equation}
    \exp(-\frac{1}{2}\lVert \bh(\bx) \rVert_{\bSigma_{obs}^{-1}}^2) = \Pi_{i=0}^N f^{obs}_i(x_i),
    \label{eq:fact_likelihood}
\end{equation}
where each factor
\begin{equation}
f^{obs}_i(x_i) \triangleq \exp(-\frac{1}{2}\lVert \bh(x_i) \rVert_{\bSigma_{obs}^{-1}}^2)
\end{equation}
represents the collision cost evaluated at corresponding support state. The collision checking needs to be carried out at a very dense set of points along the trajectory. Gaussian process representation has the advantage that the intermediate collision-checking between the support states can be done through interpolation \cite{mukadam2018continuous}, which keeps the sparsity of the representation. 
The assumptions in \eqref{eq:fact_prior} and \eqref{eq:fact_likelihood} together with the GP interpolation bring a sparse parameterization to our problem formulation and is greatly beneficial to the scalability of the proposed algorithm.

Finally, we remark that the MAP formulation \eqref{eq:MAP} can be viewed as a discretization of the following trajectory optimization
\begin{subequations}\label{eq:MAPcontrol}
\begin{eqnarray}\nonumber
    \min_{\bx(\cdot), \bu(\cdot)}\!\!\!\!\!\!\!&& \int_{t_0}^{t_N} [\frac{1}{2}\|\bu(t)\|_{\bQ_c^{-1}}^2 + \lVert \bh(\bx(t)) \rVert_{\bSigma_{obs}^{-1}}^2]dt 
    \\&&\hspace{-0.5cm}+\frac{1}{2}\lVert \bx(t_0) -\! \mu_0 \rVert_{\mathbf{K}_0^{-1}}+ \frac{1}{2}\lVert \bx(t_N) -\! \mu_N \rVert_{\mathbf{K}_N^{-1}}
    \\
    &&\dot \bx(t) = \bA(t)\bx(t) + \bF(t)\bu(t) + b(t).
\end{eqnarray}
\end{subequations}
To see this, note that, if we only evaluate $\bh(\bx)$ at discretized time $\bt = [t_0, \dots,t_N]$, then for a given $\bx = [x_0, \dots, x_N]^T$, the optimization over $\bu(\cdot)$ is a linear quadratic control problem for each time interval $(t_i, t_{i+1})$ and the corresponding closed-form minimum is exactly the exponent of $f^i_{gp}$.

\subsection{Gaussian variational inference for motion planing}
Though \eqref{eq:MAP} is a probabilistic inference problem, the solution obtained in GPMP is still deterministic in the sense that it searches for a trajectory which maximizes the posterior probability. To better capture the uncertainties and risk presented in motion planning \cite{lambert2021entropy}, we instead propose to approximate the full posterior distribution $p(\bx|\bz)$ in \eqref{eq:MAP}. 
In particular, we propose the Gaussian variational inference approach to motion planning that seeks to minimize the distance between a Gaussian distribution and the true posterior, measured by KL divergence. It reads
\begin{equation}
\begin{split}
    q^{\star} &= \underset{q\in \mathcal{Q}}{\arg\min}\; {\rm KL} [q(\bx) || p(\bx|\bz)]\\
    &= \underset{q\in \mathcal{Q}}{\arg\min}\; \mE_q{[\log q(\bx)-\log p(\bz|\bx)-\log p(\bx)]}\\
    &= \underset{q\in \mathcal{Q}}{\arg\max}\; \mE_q{\log p(\bz|\bx)} - {\rm KL} [q(\bx)||p(\bx)]
\end{split}
\label{eq:KL_min}
\end{equation}
where $\mathcal{Q}$ denotes the Gaussian distribution family. The expression $\mE_q{\log p(\bz|\bx)} - {\rm KL} [q(\bx)||p(\bx)]$ is known as the evidence lower bound (ELBO). 
The optimal distribution $q^{\star}$
encourages putting mass on the likelihood $p(\bz|\bx)$ while minimizing its distance from the prior $p(\bx)$. It shows the trade-off between the smoothness and the collision avoidance. 

An alternative form of \eqref{eq:KL_min} is
\begin{equation}
    \begin{split}
    q^{\star} &= \underset{q \in \mathcal{Q}}{\arg\max}\; \mE_q[\log p(\bx|\bz) - \log q(\bx)]\\
    &= \underset{q \in \mathcal{Q}}{\arg\max}\; \mE_q[\log p(\bx|\bz)] + H(q)
    \end{split}
    \label{eq:entropy}
\end{equation}
where $H(q) = -\mE_q[\log(q)]$ is the entropy of the distribution. The objective can thus be interpreted as Gaussian process motion planning with an entropy regularization term. 

To further balance the trade-off between the original prior-collision cost and the entropy cost, a temperature $T$ can be introduced, pointing to
\begin{equation}
\begin{split}
    q^{\star} &= \underset{q \in \mathcal{Q}}{\arg\max}\; \mE_q[\log p(\bx|\bz)] + T H(q)\\
        &= \underset{q \in \mathcal{Q}}{\arg\max}\; \frac{1}{T}\mE_q[\log p(\bx|\bz)] + H(q).
\end{split}
   \label{eq:entropy_temperature}
\end{equation} 
When the temperature is low (small $T$), the optimization puts more weight on maintaining smoothness while avoiding obstacles. When the temperature is high, more weights are put on the system entropy cost to find solutions which have larger covariances so that they can tolerate larger uncertainties.
\begin{remark}\label{remark_T}
Formulation \eqref{eq:entropy_temperature} shows an interpolation from the deterministic smooth-collision-avoiding objective \eqref{eq:MAP} to an entropy regularized robust motion planning by changing the temperature $T$. To recover the deterministic solutions, as $T$ approaches to $0$, it can be shown \cite{HazSha10} that obtained optimal value will tend to the minimal value for the original objective \eqref{eq:MAP}. Indeed, when $T \to 0$, the objective in \eqref{eq:entropy_temperature} approaches $\mE_q[\log p(\bx|\bz)]$ with respect to $q\sim \mathcal{N}(\bmu, \bSigma)$. In this case, when $\bSigma$ shrinks to 0, the objective function  $\mE_q[\log p(\bx|\bz)]$ tends to $\log p(\bmu|\bz)$.
\end{remark}

Finally, we note that the variational inference formulation \eqref{eq:entropy_temperature} can be viewed as a time discretization of the following stochastic control problem
\begin{subequations}\label{eq:VIcontrol}
\begin{eqnarray}\nonumber
    \min_{\bx(\cdot), \bu(\cdot)}\!\!\!\!\!\!\!\!\!&& \mE\{\int_{t_0}^{t_N} [\frac{1}{2}\|\bu(t)\|_{\bQ_c^{-1}}^2 + \lVert \bh(\bx(t)) \rVert_{\bSigma_{obs}^{-1}}^2]dt 
    \\&&\hspace{-0.6cm}+\frac{1}{2}\lVert \bx(t_0) -\! \mu_0 \rVert_{\mathbf{K}_0^{-1}}\!+\! \frac{1}{2}\lVert \bx(t_N) -\! \mu_N \rVert_{\mathbf{K}_N^{-1}}\}
    \\
    &&\hspace{-0.6cm} \dot \bx(t) =\! \bA(t)\bx(t) \!+\! \bF(t)(\bu(t)+T \bw(t))\! +\! b(t).
\end{eqnarray}
\end{subequations}
The proof is based on an equivalence relation between the quadratic control energy and the KL divergence ${\rm KL}(q \| p)$ \cite{CheGeoPav16}.
The only difference between \eqref{eq:VIcontrol} and \eqref{eq:MAPcontrol} is that the dynamics in \eqref{eq:VIcontrol} is disturbed by white noise $T \bw(t)$. Thus, as $T$ goes to zero, \eqref{eq:VIcontrol} should converge to \eqref{eq:MAPcontrol}.

\section{optimization scheme}
\label{sec:optimization}
GVI formulates the motion planing problem as an optimization over Gaussian distributions $q(\bx)\sim \calN (\bmu, \bSigma)$.
Denote the concatenation of the mean and covariance in vector form as $\balpha \triangleq (\bmu, vec(\bSigma^{-1}))$. The inference objective then reads
\begin{equation}
    V(q) = {\rm KL}[q(\bx)||p(\bx|\bz)] = \mE_q[\log q(\bx) - \log p(\bx|\bz)].
\label{eq:objective}
\end{equation}
To solve this optimization, we utilize the natural gradient descent scheme. The factorized objective assumption which leads to a sparsity pattern of the problem is also leveraged to improve the scalability of our algorithm.
\subsection{Natural gradient descent} 
For notation simplification, we denote $\psi(\bx) = -\log p(\bx|\bz)$. The derivatives w.r.t. $\bmu$ and $\bSigma^{-1}$ can be derived \cite{opper2009variational} explicitly
\begin{subequations}
    \begin{align}
        \frac{\partial V(q)}{\partial \bmu} &= \bSigma^{-1}\mE[(\bx-\bmu)\psi(\bx)]
        \label{eq:deriv1}\\
        \frac{\partial^2V(q)}{\partial \bmu \partial \bmu^T} &= \bSigma^{-1}\mE[(\bx-\!\bmu)(\bx\!-\!\bmu^T)\psi(\bx)]\bSigma^{-1} \!- \!\!\bSigma^{-1}\mE[\psi(\bx)]
        \label{eq:deriv2}\\
        \frac{\partial V(q)}{\partial \bSigma^{-1}} &= \frac{1}{2}\bSigma\mE[\psi(\bx)]\!-\!\frac{1}{2} \mE[(\bx-\!\bmu)(\bx-\!\bmu)^T\psi(\bx)]  + \frac{1}{2}\bSigma.
        \label{eq:deriv3}
    \end{align}
    \label{eq:derivs}
\end{subequations}
All expectations are taken w.r.t. $q$. Comparing \eqref{eq:deriv2} and \eqref{eq:deriv3} we obtain
\begin{equation}
\frac{\partial^2V(q)}{\partial \bmu \partial \bmu^T} = \bSigma^{-1} - 2\bSigma^{-1}\frac{\partial V(q)}{\partial \bSigma^{-1}} \bSigma^{-1}.
\label{eq:equiv_deriv}
\end{equation}

Having the relations in \eqref{eq:derivs} and \eqref{eq:equiv_deriv}, for Gaussian distribution $q\sim\mathcal{N}(\bmu, \bSigma)$, a natural gradient descent update step w.r.t. objective function $V$ can be calculated straightforward \cite{magnus2019matrix} as 

\begin{equation}
    \begin{bmatrix}
    \delta \bmu\\
    vec(\delta \bSigma^{-1})
    \end{bmatrix}
    = -\begin{bmatrix}
    \bSigma & \mathbf{0} \\
    \mathbf{0} & 2(\bSigma^{-1} \otimes \bSigma^{-1})
    \end{bmatrix}
    \begin{bmatrix}
    \frac{\partial V}{\partial \bmu^T}\\
    vec(\frac{\partial V}{\partial \bSigma^{-1}})
    \end{bmatrix}.
\end{equation}
Using properties of the kronecker product and vectorizations of matrices, the update step in natural gradient is
\begin{equation}
    \bSigma^{-1} \delta \bmu = -\frac{\partial V}{\partial \bmu},\;\;\; \delta \bSigma^{-1} = -2\bSigma^{-1}\frac{\partial V}{\partial\bSigma^{-1}}\bSigma^{-1}.
\label{eq:ngd_step}
\end{equation}
Notice that we write \eqref{eq:ngd_step} in terms of $\bSigma^{-1}$ to fully leverage its sparsity pattern. Comparing \eqref{eq:derivs} and \eqref{eq:ngd_step}, we have
\begin{equation}
    \begin{aligned}
    \delta \bSigma^{-1} &= \frac{\partial^2V(q)}{\partial \bmu \partial \bmu^T} - \bSigma^{-1}.
\end{aligned}
\label{eq:deltaSigma}
\end{equation}
Equation \eqref{eq:ngd_step} and \eqref{eq:deltaSigma} tells that, to calculate the update $\delta \bmu, \delta \bSigma^{-1}$, we only need to compute \eqref{eq:deriv1} and \eqref{eq:deriv2}. The new variables are calculated using the updates, a step size $\gamma < 1$, and a constant $R$ in a backtracking fashion as
\begin{equation}
    \bmu \leftarrow \bmu + \gamma^R \times \delta \bmu, \;\;\;  \bSigma^{-1} \leftarrow \bSigma^{-1} + \gamma^R \times \delta \bSigma^{-1},
\label{eq:backtracking}
\end{equation}
where $R>1$ is increasing to shrink the step size for backtracking until the cost decreases. Line search algorithms \cite{wright1999numerical} can also be deployed to obtain locally minimum solutions for this non-convex optimization.

\subsection{Factorized objectives}
We next show that with factorized cost functions, the update step in the algorithm will preserve the sparsity pattern of $\bSigma^{-1}$. Under the factorized assumptions \eqref{eq:fact_prior} and \eqref{eq:fact_likelihood}, and denote $\psi_k(\bx_k) = -\log p(\bx_k|\bz)$, \eqref{eq:objective} also factorizes
\begin{equation}
    \begin{split}
        V(q) &= \mE_q[\log q(\bx)] - \sum_{k=1}^K \mE_{q_k}[\psi_k(\bx_k)]\\
        &= \mE_q[\log q(\bx)] - \sum_{k=1}^K \mE_{q_k}[\log p(\bx_k) + \log p(\bz|\bx_k)]\\
        &\triangleq \frac{1}{2} \log (\lvert \bSigma^{-1} \rvert) + \sum_{k=1}^K V_k(q_k)
    \end{split}
    \label{eq:factorized_obj}
\end{equation}
where $V_k(q_k)$'s are factored costs and $\bx_k$ are the corresponding subsets of variables to the $k$th factor. We assume that $\bx_k$ can be transformed from $\bx$ using a linear mapping $\bM_k$, i.e., $\bx_k = \bM_k\bx$, and the marginal Gaussian $q_k \sim \calN(\bmu_{k}, \bSigma_{k})$. The relation between the joint and the factorized variables reads
\begin{equation}
    \bmu_{k} = \bM_k \bmu,\;\;\; \bSigma_{k} = \bM_k \bSigma \bM_k^T.
    \label{eq:transform_factor}
\end{equation}
In view of \eqref{eq:derivs} and \eqref{eq:ngd_step}, to compute the updates $\delta \bmu$ and $\delta \bSigma^{-1}$, we need to calculate the derivatives of the joint objective which also factorizes as
\begin{subequations}
\begin{align}
    \frac{\partial V(q)}{\partial \bmu} &= \sum_{k=1}^K \bM_k^T \frac{\partial V_k(q_k)}{\partial \bmu_{k}},\\
    \frac{\partial^2 V(q)}{\partial \bmu \partial \bmu^T} &= \sum_{k=1}^K \bM_k^T \frac{\partial^2 V_k(q_k)}{\partial \bmu_{k} \partial\bmu_{k}^T} \bM_k.
\end{align}
\label{eq:fact_update}
\end{subequations}
The factorized derivatives $\frac{\partial V_k}{\partial \bmu_{k}}$ and $\frac{\partial^2 V_k}{\partial \bmu_{k} \bmu_{k}^T}$ will have the same expressions  as in \eqref{eq:derivs} w.r.t. marginal distributions $q_k\sim \calN(\bmu_k, \bSigma_k)$ and marginal factors $\psi(\bx_k)$
\begin{subequations}
    \begin{align}
        \frac{\partial V_k}{\partial \bmu_{k}} &= \bSigma_{k}^{-1}\mE_{q_k}[(\bx_k-\bmu_{k})\psi(\bx_k)],\label{eq:deriv1_k}\\
        \frac{\partial^2V_k}{\partial \bmu_{k} \partial \bmu_{k}^T} &= \bSigma_{k}^{-1}\mE_{q_k}[(\bx_k-\bmu_{k})(\bx_k-\bmu_{k}^T)\psi(\bx_k)]\bSigma_{k}^{-1}\nonumber\\
        & - \bSigma_{k}^{-1}\mE_{q_k}[\psi(\bx_k)].\label{eq:deriv2_k}
    \end{align}
    \label{eq:derivs_k}
\end{subequations}
From \eqref{eq:deltaSigma}, \eqref{eq:fact_update} and  \eqref{eq:derivs_k} we see that the sparsity pattern of the precision matrix $\bSigma^{-1}$ is preserved after the transitions between the joint and factorized updates. 

From \eqref{eq:transform_factor} we know that a joint covariance matrix $\bSigma$ is computed in each update step. Throughout the iterations $\bSigma^{-1}$ remains sparse, but $\bSigma$ need not to be. However, because of the consistent sparsity pattern, efficient methods \cite{broussolle1978state} exist in sparse linear algebra literature to compute only the parts of $\bSigma$ corresponding to the non-zero elements in $\bSigma^{-1}$. Alternatively, Gaussian belief propagation \cite{bickson2008gaussian} \cite{Ortiz2021visualGBP} can also solve the marginal covariance efficiently. The expectations in \eqref{eq:derivs_k} are approximately evaluated using Gauss-Hermite quadrature \cite{arasaratnam2007discrete} in this work. We note that when the posterior $p(\bx|\bz)$ is linear, then expectations in \eqref{eq:derivs} have closed-form, which greatly accelerates the algorithm.

\section{experiments}
\label{sec:experiments}
In all our experiments, we consider a constant-velocity model in \eqref{eq:dynamics}. Let 
\begin{equation}
    \bA(t) = \begin{bmatrix}
    \mathbf{0} & \bI \\
    \mathbf{0} & \mathbf{0}
    \end{bmatrix},~ 
    \bb(t)=\begin{bmatrix}
        \mathbf{0}\\
        \mathbf{0}
    \end{bmatrix},~ 
    \bF(t)=\begin{bmatrix}
    \mathbf{0} \\ \bI
    \end{bmatrix}.
\end{equation}
The transition matrix $\bPhi$, matrices $\bQ_i$, $\bQ_i^{-1}$, $\bQ$, and $\bB$ in \eqref{eq:tehta_inv_A} can be calculated explicitly \cite{barfoot2014batch}. The likelihood function is defined the same as in \cite{mukadam2018continuous} \cite{ratliff2009chomp} by
\begin{equation}
    \bh(\bx) = \bc_{\epsilon}(\bd(FK(\bx)))
\end{equation}
where $FK(\cdot)$ is the forward kinematics, $\bd(\cdot)$ is the signed distance function given a signed distance field (SDF), and $\bc_{\epsilon}(\cdot)$ is the hinge loss function 
\begin{equation}
\bc_{\epsilon}(y) = 
\begin{cases} 
      0, & \text{if } y \geq \epsilon\\
        \epsilon - y, & \text{if } y < \epsilon.
\end{cases}
\label{eq:hinge_loss}
\end{equation}
When evaluating the signed distance function $\bd(\cdot)$, robots are modeled as balls with fixed radius $r$ \cite{mukadam2018continuous} at designated locations. The minimum distance from robots to obstacles is efficiently computed using the distance between centers of the balls to the obstacles and the ball radius. In this paper, to highlight the convergence of the algorithm, GP interpolation is not involved in any experiments.
\subsection{2d point robot collision avoidance}
\begin{figure}[ht]
    \centering
    \includegraphics[width=0.99\linewidth]{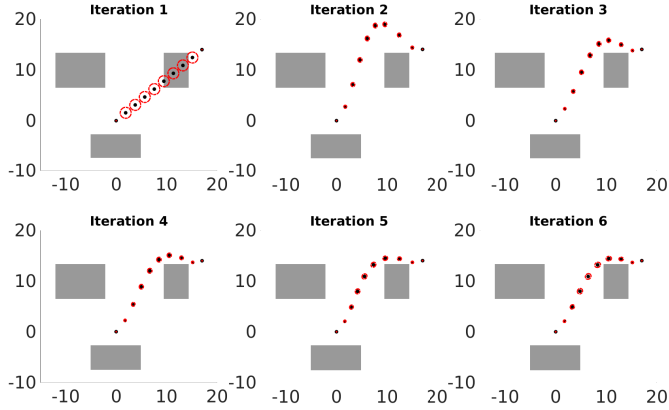}
    \caption{Converging process with $T=10, \bQ_c = 0.8\bI, \bSigma_{obs} = 0.004\bI, \epsilon = 0.7.$ Linear interpolated initialization for $\bmu$ and $10 \bI$ for $\bSigma^{-1}$.}
    \label{fig:planar_pR_conv_process_above}
\end{figure}

The first experiment is conducted with a planar point robot, which better captures the idea of covariance by plotting ellipsoids. Fig.\ref{fig:planar_pR_conv_process_above} shows the convergence of the support states. Black dots represent $\bmu$, and the red ellipsoids draw the $0.997$ confidence region contour. We initialize $\bmu$ using a linear interpolation between the start and goal states, and initialize $\bSigma^{-1}$ using isotropic matrices. 

\begin{figure}[ht]
    \centering
    \includegraphics[width=0.99\linewidth]{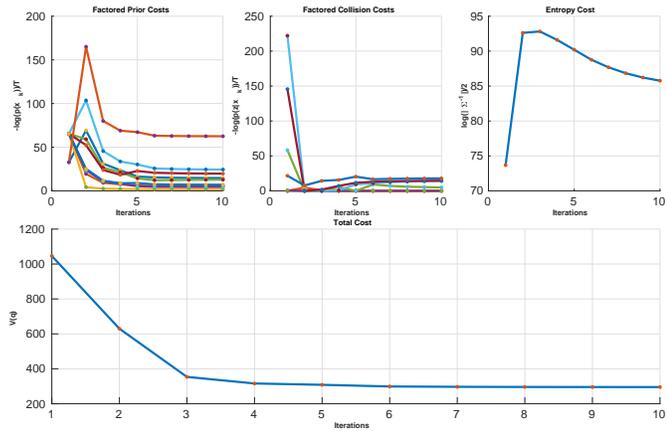}
    \caption{Decomposed and total costs. Prior and collision costs on the factor level and the entropy cost on the joint level.}
    \label{fig:planar_pR_loss_above}
\end{figure}

\paragraph{Trade-off between motion planning and system entropy}
The cost function in \eqref{eq:factorized_obj} contains two parts: a motion planning including prior and collision costs, and a regularized entropy cost. Fig. \ref{fig:planar_pR_loss_above} shows the evolution of different costs and the total cost, where the prior and collision costs are factorized, and the cost on the entropy $\frac{1}{2}\log(\lvert \bSigma^{-1} \rvert)$ is computed on the joint level. As shown in Fig.\ref{fig:planar_pR_loss_above}, during the first several iterations the prior and collision costs on each factor decreases, meaning that the system gets rid of the obstacle while maximizing trajectory smoothness and system dynamics assumptions imposed by the prior. Meanwhile, the entropy costs increase. After the system is safe and smooth, the algorithm moves to the region where the entropy cost decreases. During the two phases, the total loss decreases. This trade-off process is also reflected in the Fig. \ref{fig:planar_pR_conv_process_above}. The covariance pivots shrink while the system is avoiding the obstacles, and increase after the system is safe and smooth.



\begin{figure}
    \centering
    \includegraphics[width=0.99\linewidth]{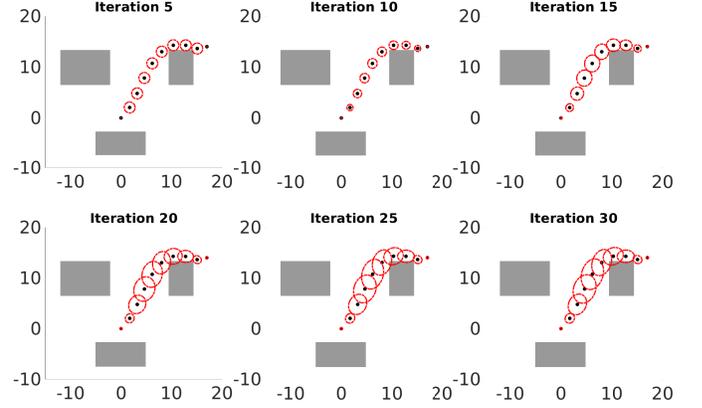}
    \caption{Convergence with $T = 100$, using means from the last iteration in Fig. \ref{fig:planar_pR_conv_process_above} as initialization for $\bmu$. Other parameters are same as in Fig. \ref{fig:planar_pR_conv_process_above}.}
    \label{fig:high_T_above_conv}
\end{figure}

\paragraph{Planning with high temperature}
In \eqref{eq:entropy_temperature}, a temperature $T$ is introduced to alter the weights between planning objective and entropy cost. To achieve feasible trajectories, we use small $T$. However in low temperature regions, little changes on $\bSigma$ will happen due to the low weight on the entropy cost. One motivation of the proposed formulation is that we would like to leverage the entropy in order to have wider-spread distributions in all areas, since the $3\sigma$ area measures the size of the safe regions in a probabilistic sense. Higher temperature promotes the system's entropy, but put less weights on the feasibility part. A compromise is to use a near-feasible initialization with high temperature. The initialization for the mean $\bmu$ can either be the output of a lower temperature optimization as a re-planning, or from a higher level sampling based planner. Fig. \ref{fig:high_T_above_conv} shows the converging process of the iterations for a high temperature re-planning. We note that the low temperature planning and the high temperature re-planning can be done in a consecutive manner in the optimization.

\subsection{More challenging planning problems}
\begin{figure}
    \centering
    \includegraphics[width=0.99\linewidth]{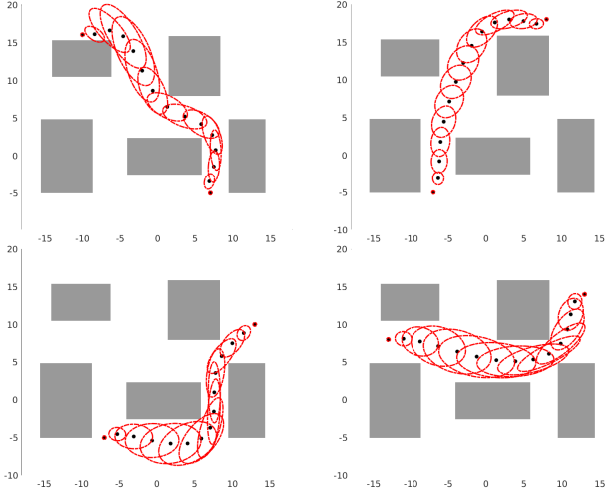}
    \caption{Planning in cluttered environments. $15$ support states, $\bQ_c = 0.8\bI, \bSigma_{obs} = 0.0035\bI \sim 0.0045\bI, r=1.5,  \epsilon = 0.7.$ All plans \textit{`go-through`} a low temperature planning and a high temperature re-planning.}
    \label{fig:planning_cluttered}
\end{figure}

In the next set of experiments we show that by introducing entropy regularization to the deterministic formulation, we gain flexibility in solution searching as well as a risk-measuring metric. We illustrate using several experiments. In paragraph (a), to test the performance in hard tasks, we conduct long range planning in cluttered environments. In (b) we use a narrow gap environment to show that stochasticity brings flexibility in choosing collision-checking radius, compared with deterministic baseline; In (c) we show that stochasticity help explore solution spaces and find multiple locally optimal candidate solutions. In (d) it is shown that entropy serves as a measure of risk which plays an important role in decision making in terms of choosing the final plan.
\paragraph{Long distance planning in cluttered environments}\label{paragraph:a_long_dist}
We first conduct long-distance tasks in a cluttered environment for a planar point robot. Fig.\ref{fig:planning_cluttered} shows the resulting trajectory distributions. In practice we found that the smoothness captured by Gaussian processes is the key for the trajectories to circumvent sharp corners and achieve long distance targets. We observe that the covariances shrink in the narrow areas and stretch in the safe zones. The volume of the confidence regions describes level of safety locally, since when sampling trajectories from the distributions, regions with wider confidence region provides more choices with the same level of confidence on feasibility. The adaptive confidence regions brings robustness to the trajectories in face of environment uncertainties.

\begin{figure}[ht]
    \centering
    \includegraphics[width=0.99\linewidth]{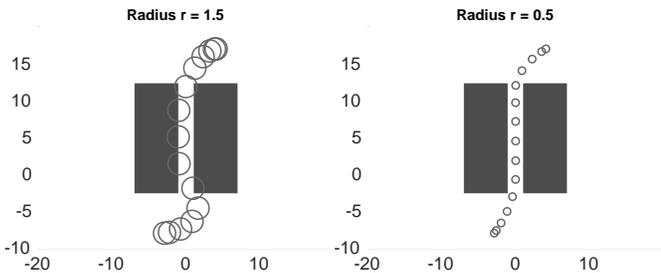}
    \caption{Results of GPMP2 for different collision-checking radius. $15$ support states. $\bQ_c = 0.8\bI, \bSigma_{obs} = 0.0055\bI, \epsilon = 0.6$. Linear interpolated initialization for both figures.}
    \label{fig:gpmp2_comparing_radius}
\end{figure}

\begin{figure}
    \centering
    \includegraphics[width=0.99\linewidth]{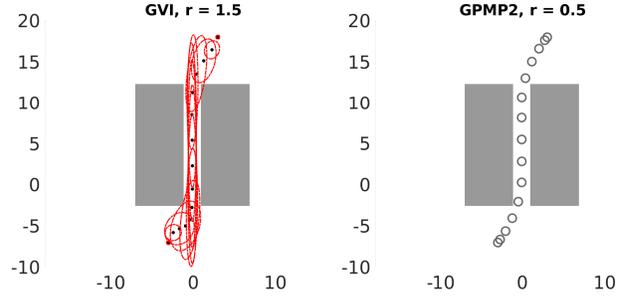}
    \caption{Comparison of the stochastic and deterministic \textit{`go-through'} plan.  $r=1.5$ for GVI, and $r=0.5$ for the GPMP2. Other parameters are the same as in Fig. \ref{fig:gpmp2_comparing_radius}. Note that the same parameters lead to a successful plan in the left figure, compared with the left figure in Fig. \ref{fig:gpmp2_comparing_radius}.}
    \label{fig:comparison_narrow_gpmp2}
\end{figure}

\paragraph{Planning through a narrow gap with more flexible collision-checking radius}
Fig. \ref{fig:gpmp2_comparing_radius} shows the planning task in a narrow-opening environment. We first show that the covariance can provide flexibility in collision checking. For the deterministic baseline GPMP2, the radius $r$ of collision-checking balls needs to be prefixed and in accordance with the environment. Fig. \ref{fig:gpmp2_comparing_radius} shows that $r$ needs to be small enough to achieve a successful \textit{`go-through'} plan. In Fig. \ref{fig:comparison_narrow_gpmp2}, our proposed method can obtain a successful motion plan using the same radius which has led to a failed plan in GPMP2 shown in the left subfigure in Fig. \ref{fig:gpmp2_comparing_radius}. We note that this is because that the proposed method optimizes directly over covariance so that the expected cost \eqref{eq:factorized_obj} can always decrease even with large collision-checking radius. In complex planning tasks, variable covariance can give flexibility in choosing $r$ as one hyperparameter. In real-world  planning tasks, different levels of safety are required in different regions in the environment, which is directly encoded in the variable covariance.

\begin{figure}[ht]
    \centering
    \includegraphics[width=0.99\linewidth]{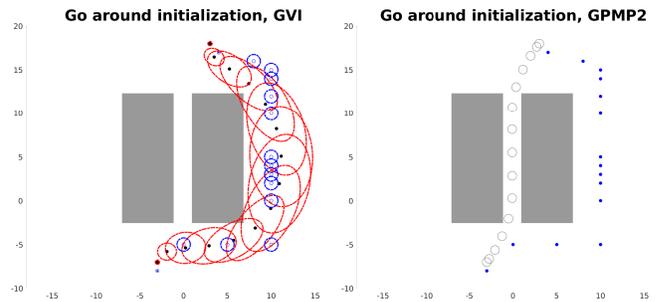}
    \caption{Solution space exploration comparison. $15$ Supported stats,  $\bQ_c = 0.8\bI, \bSigma_{obs} = 0.0055\bI, \epsilon = 0.6$, $r=1.5$ for the left, and $r=0.5$ for the right figure \protect \footnotemark. Both figures use a same \textit{`go-around`} course initialization as shown in blue.}
    \label{fig:gpmp2_compare_initialization}
\end{figure}

\paragraph{Plan circumventing a narrow gap showing solution space exploration}
We show by experiment that the entropy regularization can also promote solution space exploration. Trajectory optimization is often initialized using a sampling-based course plan such as RRT \cite{lavalle1998rapidly} \cite{tedrake2010lqr, majumdar2017funnel}, which is partially because that the problem is non-convex and it is easier to find a local optimal value if started closer. In Fig. \ref{fig:gpmp2_compare_initialization}, a \textit{`go-around`} initialization is used for both the proposed method and GPMP2, other parameters being the same. Starting from the same initialized seed, the proposed framework finds a \textit{`go-around'} trajectory circumventing the gap while GPMP2 converged back to the \textit{`go-through'} plan. This shows that stochasticity encourage solution domain exploration in finding candidate motion plans. As explained in the next paragraphs, this is because the entropy cost regularizes the total cost.

\footnotetext{$r=1.5$ is also tested for the GPMP2, which converges to the narrow gap with collisions.}

\begin{figure}[ht]
    \centering
    \includegraphics[width=0.99\linewidth]{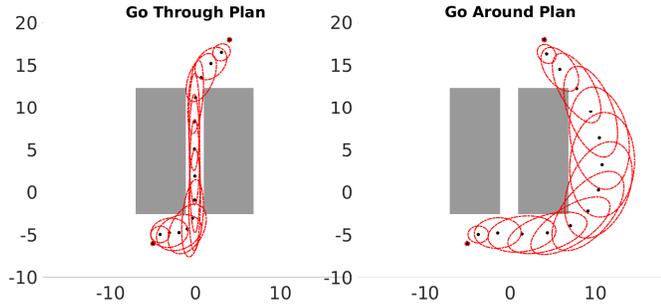}
    \caption{An example of comparing different motion plans. Entropy regularizes the motion planning objective.  Costs are shown in Tab. ???\ref{tab:narrow_comparison}.}
    \label{fig:comparison_narrow}
\end{figure}

\paragraph{Comparing locally minimum solutions leveraging entropy}
When comparing different solutions, the entropy cost serves as a risk-measuring metric in addition to motion planning costs. Intuitively, plans with lower entropy cost are considered to be less risky, because the covariance stretches wider in safer regions. As an example, Fig. \ref{fig:comparison_narrow} compares two motion plans visually, and Tab. \ref{tab:narrow_comparison} compares different costs for the two plans in Fig. \ref{fig:comparison_narrow}. Results show that the \textit{`go-around'} plan has far lower collision and entropy costs which together beat the \textit{`go-through'} plan. In this scenario, it is reasonable to choose a longer but less risky \textit{`go-around'} plan which circumvents the narrow gap.

\begin{table}[ht]
\centering
\begin{tabular}{|c|c|c|c|c|c|}
\hline
 & \textbf{Prior} & \textbf{Collision} & \textbf{MP} & \textbf{Entropy} & \textbf{Total} \\ \hline
Left & \textbf{34.4583} & 9.1584 & \textbf{43.6168}  & 44.1752 & 87.7920 \\ \hline
Right & 42.9730 & \textbf{2.0464} & 45.0193 & \textbf{39.9193} & \textbf{84.9387} \\ \hline
\end{tabular}
\caption{Comparing costs for plans in Fig. \ref{fig:comparison_narrow}. The regularized entropy cost $\log(\lvert\bSigma^{-1}\rvert)$ changed the order of the total costs and enables the optimization to choose a probabilistically less risky solution. \textit{`MP`} stands for motion planning costs which is the sum of prior and collision costs.}
\label{tab:narrow_comparison}
\end{table}

\subsection{Arm robot}
\begin{figure}[ht]
    \centering
    \includegraphics[width=0.99\linewidth]{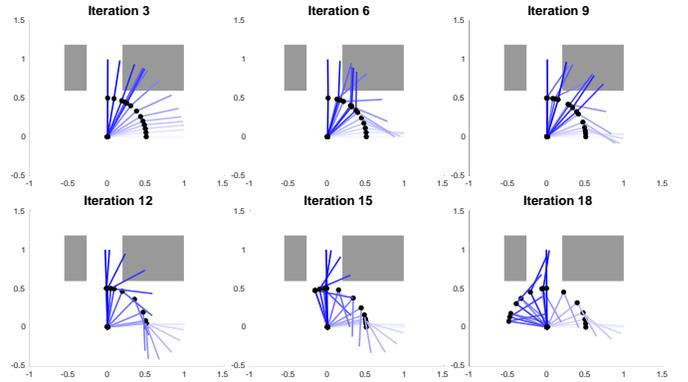}
    \caption{Convergence for 2-link arm in a cluttered environment.}

    \label{fig:arm_map2_iterations}
\end{figure}

\begin{figure}
    \centering
    \includegraphics[width=0.99\linewidth]{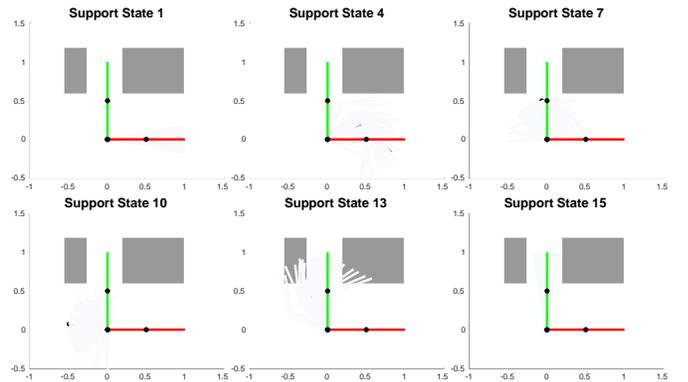}
    \caption{Motion plan for a 2-link arm. Red state is the start, green is the goal, and solid blue states are selected means of support states, and shadowed blue states are samples.}
    \label{fig:2d_arm_samples}
\end{figure}

To validate our proposed framework, we conducted experiments on a 2 types of arm robots.
\paragraph{2-link arm model}
Fig. \ref{fig:arm_map2_iterations} shows the convergence process in a cluttered environment. Fig. \ref{fig:2d_arm_samples} shows the sampled states from the obtained distributions. The last iteration in Fig. \ref{fig:arm_map2_iterations} shows a reasonable collision avoidance behavior while keeping the smoothness of the trajectory. In Fig. \ref{fig:2d_arm_samples}, we plot the means and samples for the support states of the last iteration in Fig. \ref{fig:arm_map2_iterations}. The solid blue bars represent the mean values, and shadowed bars are samples. The depth of the shadowed states represents the sample frequency. As shown in Fig. \ref{fig:2d_arm_samples}, in less cluttered area, samples distribute wider, representing higher entropy, and in the more constrained areas, there are less freedom.

\begin{figure}
    \centering
    \includegraphics[width=.99\linewidth]{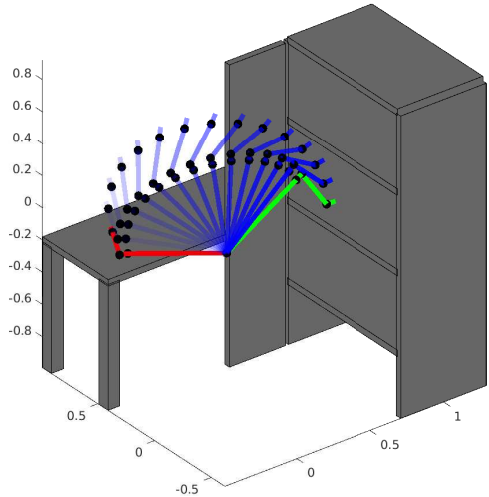}
    \caption{Mean values for the supported states.}
    \label{fig:wam_mean}
\end{figure}

\begin{figure}
    \centering
    \includegraphics[width=.99\linewidth]{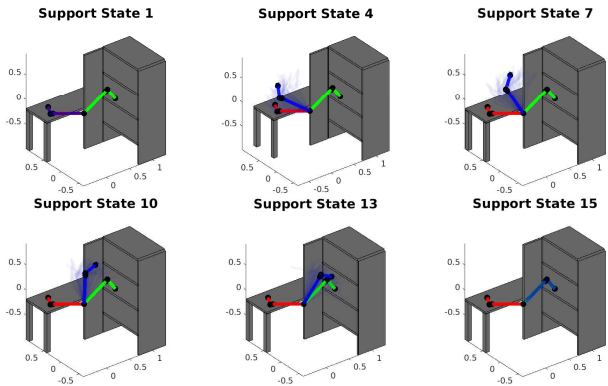}
    \caption{Mean values and samples for each supported state.}
    \label{fig:wam_samples}
\end{figure}

\paragraph{7-DOF WAM arm model}
Solving the optimization in the space of distributions brings additional computation complexities compared with the deterministic formulation. However, the factorized cost function \eqref{eq:factorized_obj} and partial update schemes \eqref{eq:derivs_k} mitigate the problem. In addition, there exist more efficient methods in evaluating the integrals in \eqref{eq:derivs_k}, which can further accelerate the algorithm. We evaluated the proposed algorithm on a 7-DOF WAM Arm robot in a more realistic dataset, the optimized mean and samples are shown in Fig. \ref{fig:wam_mean} and Fig. \ref{fig:wam_samples}. 

\section{conclusion}\label{sec:conclusion}
In this work we proposed a Gaussian variational inference framework to approach motion planning as a probability inference. On top of the Gaussian process representation of the trajectory, we calculate an optimal Gaussian distribution over the trajectories. Natural gradient descent scheme was deployed to solve the GVI. Factorized cost functions brings a sparsity pattern into the framework, and Gaussian assumption brings an explicit update scheme which converges quickly to locally minimum solutions. Alternatively, the proposed framework can be viewed as motion planning with entropy regularization. Experiments show that the proposed method achieves smooth collision-free trajectories, and also provides more robust solutions than deterministic baseline methods, especially in challenging environments. The limitation of the proposed algorithm is the  computation complexity increased by introducing additional optimization variables, which is a trade-off for the additional distributional information gain. However, this issue can be mitigated by leveraging the problem's sparsity pattern and more advanced integration estimation techniques.

\bibliographystyle{Bibliography/IEEETrans}
\bibliography{root}

\end{document}